%% file: main.tex
\documentclass[letterpaper]{article} 
\usepackage[preprint]{aaai2027}  
\usepackage[hyphens]{url}  
\usepackage{graphicx} 
\usepackage{natbib}  
\usepackage{caption} 
\usepackage{algorithm}
\usepackage{algorithmic}
\usepackage{multirow}
\usepackage{amsmath}
\usepackage{amssymb}
\usepackage{amsthm}
\usepackage{subcaption}
\usepackage{newfloat}
\usepackage{listings}
\usepackage{array} 
\newtheorem{theorem}{Theorem}
\DeclareCaptionStyle{ruled}{labelfont=normalfont,labelsep=colon,strut=off} 
\floatstyle{ruled}
\newfloat{listing}{tb}{lst}{}
\floatname{listing}{Listing}

\def\method{CReSL} 

\usepackage{booktabs}

\title{Cross-Resolution Semantic Learning for Graph Domain Adaptation}
\author{
Yingxu Wang\textsuperscript{\rm 1,*},
Haoze Huang\textsuperscript{\rm 2,*},
Zhongkai Zheng\textsuperscript{\rm 3},
Shangsong Liang\textsuperscript{\rm 4,\textdagger}
}

\affiliations{
\textsuperscript{\rm 1}Mohamed bin Zayed University of Artificial Intelligence\\
\textsuperscript{\rm 2}Fuzhou University\\
\textsuperscript{\rm 3}Sun Yat-sen University\\
\textsuperscript{\rm 4}Macao Polytechnic University\\
\textsuperscript{*}Equal contribution.
\textsuperscript{\textdagger}Corresponding author.\\
yingxv.wang@gmail.com,
132301146@fzu.edu.cn,
chengchhoi@mail2.sysu.edu.cn,
liangshangsong@gmail.com
}

\begin{document}

\maketitle

\begin{abstract}
Graph Domain Adaptation (GDA) transfers predictive knowledge from labeled source graphs to unlabeled target graphs under distribution shift. Existing methods align representations or regularize graph structures, but do not explicitly model how class-discriminative knowledge learned at different source neighborhood ranges should be routed across target ranges. We call the neighborhood range encoded by a graph representation its \emph{propagation resolution} and define \emph{semantic resolution shift} as a cross-domain change in the propagation resolutions at which class-discriminative evidence is strongest. Such shifts can make fixed same-resolution pairing suboptimal and increase the risk of negative transfer. To address this issue, we propose \textbf{C}ross-\textbf{Re}solution \textbf{S}emantic \textbf{L}earning (\method{}), a GDA method that learns soft source-to-target resolution correspondence from cross-domain class structure. First, \method{} constructs a multi-resolution representation bank using a shared Graph Neural Network and learnable resolution embeddings, with a resolution-indexed expert for each source resolution. Second, \method{} introduces Cross-Resolution Prototype Transport, which constructs class-resolution prototypes from source labels and soft target posteriors and converts cross-domain prototype discrepancies into expert-specific routing over target resolutions. Third, \method{} introduces Cross-Resolution Target Grafting, which constructs posterior-weighted target-to-source prototype displacements and enforces correspondence-weighted prediction consistency for instance-level adaptation under class uncertainty. Extensive experiments on graph benchmarks under diverse domain shifts show that \method{} outperforms strong representative baselines across most settings.
\end{abstract}

\input{code/1_intro}
\input{code/2_related_work}
\input{code/4_method}
\input{code/5_exp}

\input{code/6_conclusion}

\bibliography{reference}

\end{document}

%% file: code/1_intro.tex
\section{Introduction}
Graph Domain Adaptation (GDA) aims to transfer predictive knowledge from a labeled source graph domain to an unlabeled target graph domain under distribution shift~\cite{wu2020unsupervised,cai2024graph,wu2024graph}. By leveraging source supervision together with unlabeled target data, GDA reduces reliance on costly target annotations and facilitates the deployment of graph learning models across datasets collected from different environments, platforms, or time periods~\cite{wang2024degree,yin2023coco}.

\begin{figure}[t]
    \centering
    \includegraphics[width=\columnwidth]
    {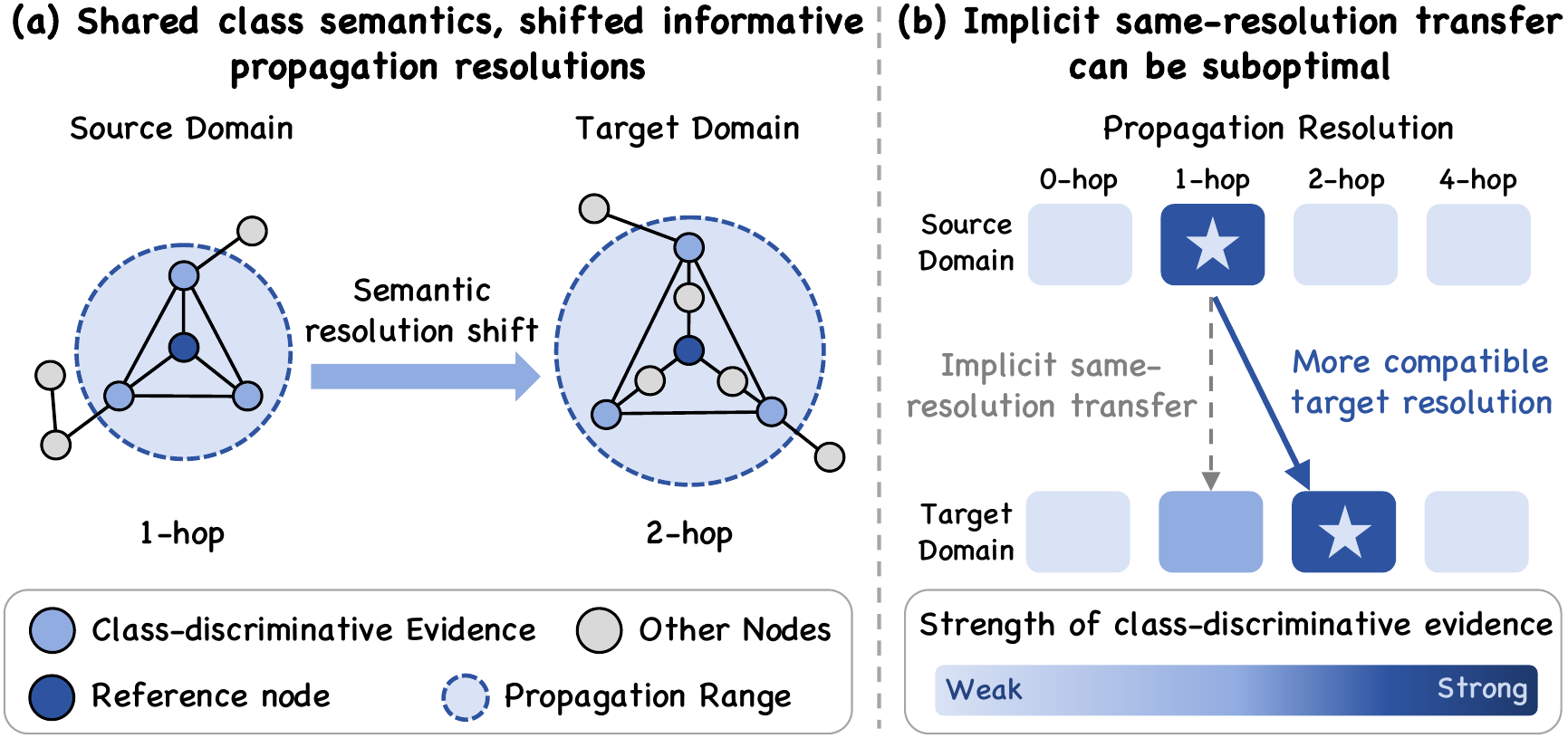}
    \vspace{-0.3cm}
    \caption{
Semantic resolution shift in GDA.
(a) Informative propagation resolutions differ across domains.
(b) Compatible source--target resolutions differ in propagation range.
}
\label{fig:semantic_resolution_shift}
\vspace{-0.3cm}
\end{figure}

Existing GDA methods improve transfer through adversarial or discrepancy-based representation alignment~\cite{wu2023non,wang2026nested}, cross-domain contrastive learning~\cite{wang2026dsbd,ma2026dual}, and pseudo-label self-training~\cite{luo2025sparse}. Graph-specific variants further exploit topology-aware reweighting~\cite{liu2023structural}, propagation calibration~\cite{liu2024rethinking}, structural consistency~\cite{you2023graph}, and class-conditioned alignment~\cite{liu2024pairwise}. However, a key limitation remains: these methods mainly align shared representations or calibrate propagation at the domain level, leaving the correspondence between source and target neighborhood ranges unspecified~\cite{chen2026adaptive,lei2025gradual,xiang2026safety}. We refer to the neighborhood range encoded by a graph representation as its \emph{propagation resolution}~\cite{pilavci2024graph}. Under domain shift, the resolution profile of class-discriminative evidence may change even when class semantics are preserved; we term this phenomenon \textbf{\emph{semantic resolution shift}}. As illustrated in Figure~\ref{fig:semantic_resolution_shift}, knowledge learned at one source resolution may therefore be more compatible with a different target resolution, making implicit same-resolution transfer prone to resolution mismatch and negative transfer~\cite{yin2025dream,xiao2025spa++}. This raises a key question: \textit{how can GDA match each source resolution to compatible target resolutions?}

\begin{figure}[t]
    \centering
    \includegraphics[width=\columnwidth]{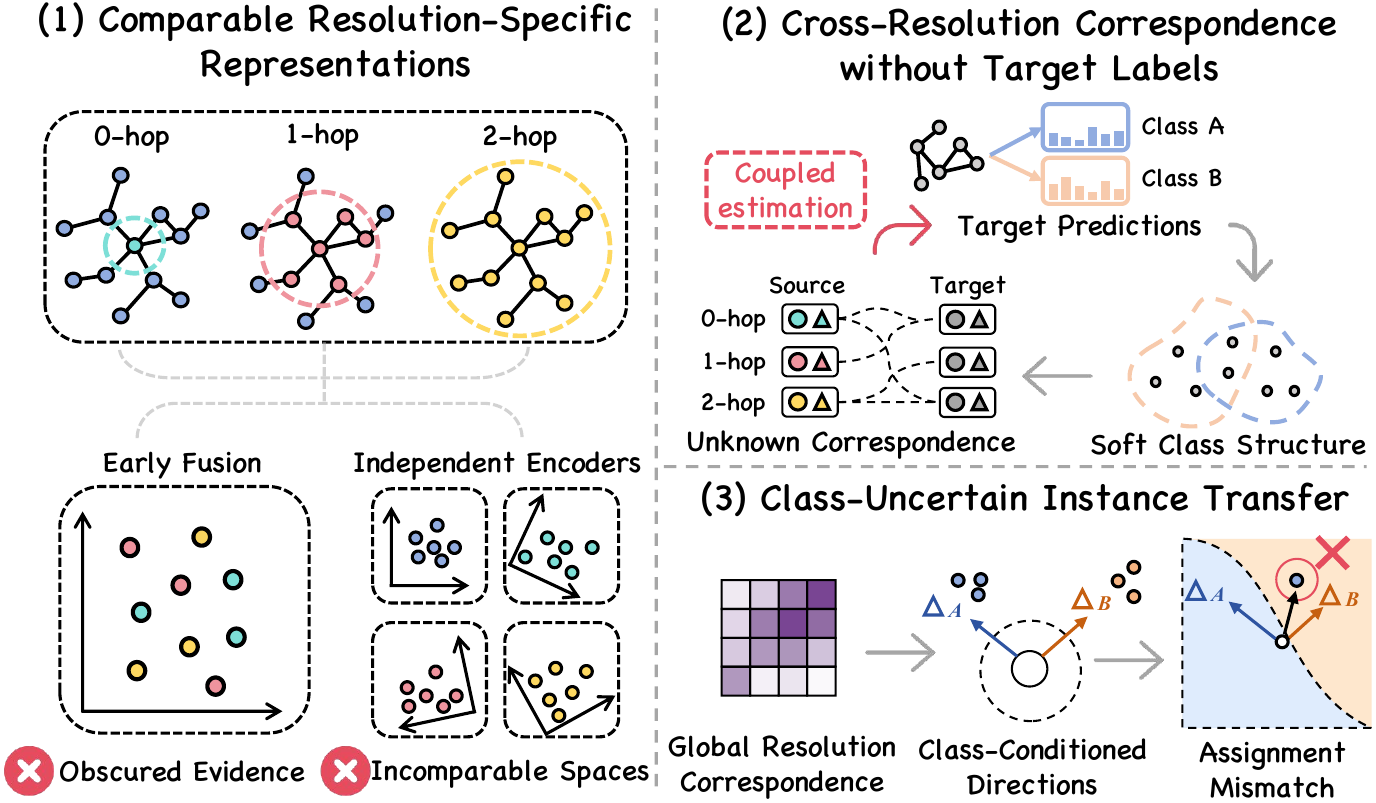}
    \vspace{-0.2cm}
    \caption{
Challenges in learning source-to-target resolution correspondence:
(1) preserving resolution-specific evidence in a comparable space;
(2) inferring correspondence under coupled target-posterior estimation;
and (3) applying global correspondence to target graphs under class uncertainty.
}
    \label{fig:challenges}
    \vspace{-0.3cm}
\end{figure}

We aim to learn a source-to-target resolution correspondence informed by cross-domain class structure. This objective raises three design challenges, as illustrated in Figure~\ref{fig:challenges}: \textbf{(1) Comparable Resolution-Specific Representations.} Representations at different propagation resolutions must preserve distinct class-discriminative evidence while remaining comparable across resolutions and domains~\cite{xu2018representation,qiao2023semi}. Direct fusion may erase resolution identity, whereas separate encoders may confound genuine resolution differences with encoder-specific variation~\cite{fang2025homophily,wang2024degree}. \textbf{(2) Cross-Resolution Correspondence without Target Labels.} Resolution compatibility is not directly observable in the target domain, while structural statistics or marginal feature similarity may fail to reflect class-conditional organization. Moreover, the target posteriors required to estimate this organization also depend on the unknown correspondence, creating a coupled estimation problem in which premature hard assignments can reinforce early errors~\cite{jiang2020implicit,wenprogressive}. \textbf{(3) Instance-Level Transfer under Class Uncertainty.} A global correspondence captures average cross-domain compatibility but does not determine how it should guide each target graph~\cite{kim2021cross,dan2024tfgda}. Since different classes may require different transfer directions and the class membership of each target graph is uncertain, hard assignment may shift its representation toward an incorrect semantic region~\cite{wang2026riemannian}.

To address these challenges, we propose \textbf{C}ross-\textbf{Re}solution \textbf{S}emantic \textbf{L}earning (\method{}), a GDA method that learns source-to-target resolution correspondence informed by cross-domain class structure, as illustrated in Figure~\ref{fig:framework}. First, \method{} constructs a multi-resolution representation bank with a shared GNN, learnable resolution embeddings, and resolution-specific experts, preserving distinct evidence in a common latent space. Second, \method{} introduces \textit{Cross-Resolution Prototype Transport} (CRPT), which jointly refines target posteriors and resolution correspondence using source and target class-resolution prototypes, and converts prototype discrepancies into expert-specific routing over target resolutions. Third, \method{} introduces \textit{Cross-Resolution Target Grafting} (CRTG), which transforms the global correspondence into target-specific adaptation by mixing class-conditioned target-to-source prototype displacements according to each target posterior and enforcing routing-weighted prediction consistency. Extensive experiments on widely used graph benchmarks under diverse domain shifts demonstrate that \method{} outperforms representative state-of-the-art baselines across most cases.

Our contributions can be summarized as follows: (1) We formulate \emph{semantic resolution shift} in GDA as the cross-domain change in the resolution profile of class-discriminative evidence. (2) We propose \method{}, featuring multi-resolution encoding with resolution-specific experts, CRPT for prototype-guided cross-resolution correspondence learning, and CRTG for posterior-conditioned instance-level target adaptation. (3) Extensive experiments on widely used graph benchmarks under diverse domain shifts demonstrate that \method{} outperforms representative state-of-the-art baselines across most settings.

%% file: code/2_related_work.tex
\section{Related Work}

\noindent\textbf{Graph Domain Adaptation.} GDA transfers predictive knowledge from a labeled source domain to an unlabeled target domain under distribution shift~\cite{wu2020unsupervised,cai2024graph}. Existing approaches mainly rely on representation alignment, target self-training, and structure-aware adaptation~\cite{liu2024revisiting,dan2024tfgda}. Adversarial and discrepancy-based objectives reduce cross-domain distribution gaps, while contrastive learning and pseudo-labeling exploit unlabeled target data~\cite{dai2022graph}. Graph-specific methods further incorporate topology reweighting, neighborhood preservation, structural regularization, or class-conditioned alignment~\cite{yin2025coupling,pang2023sa}. Recent propagation-aware approaches also adjust diffusion depth or exploit higher-order neighborhoods to mitigate structural mismatch~\cite{dan2024hogda,shi2023improving}. However, these methods typically align a shared latent space, refine target predictions, or calibrate propagation at the domain level, without explicitly modeling how knowledge learned at each source propagation resolution should correspond to target resolutions~\cite{huang2024can,chen2026adaptive}. In contrast, \method{} learns soft cross-resolution correspondence from cross-domain class-resolution prototypes and applies it through posterior-conditioned target adaptation.

\begin{figure*}[t]
    \centering
    \includegraphics[width=0.95\textwidth]{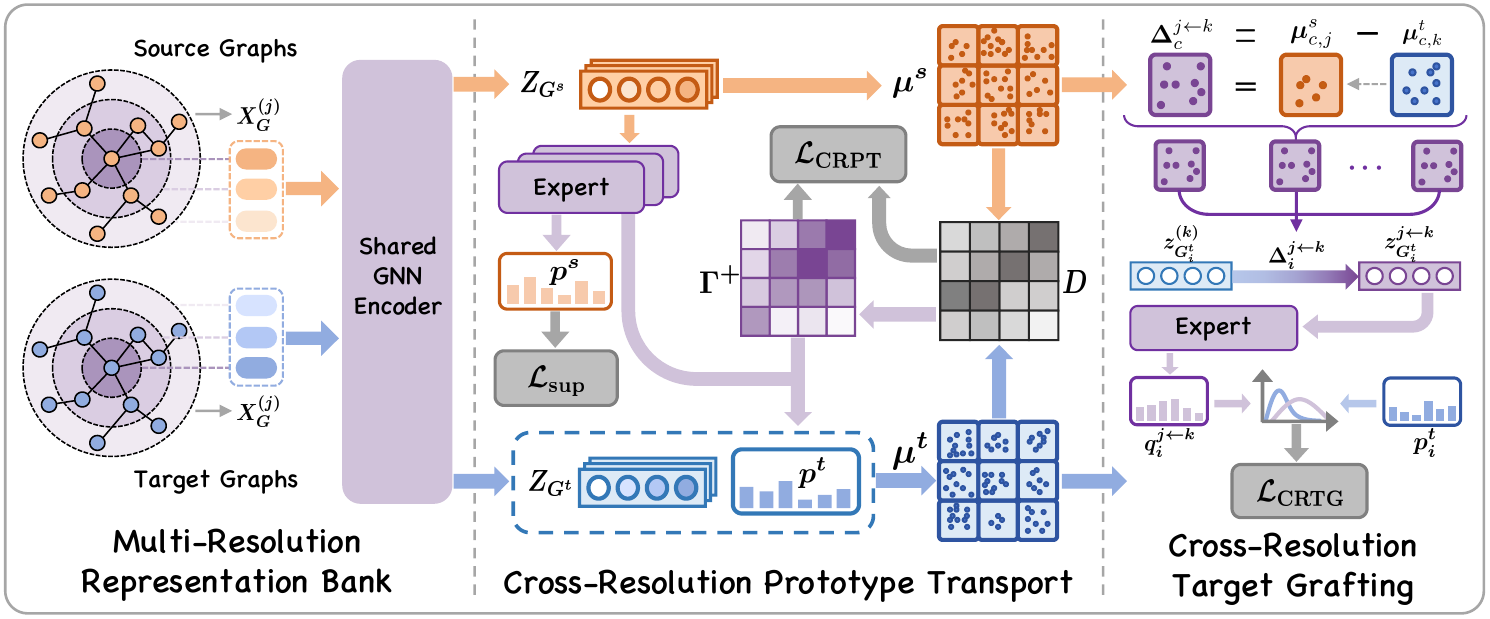}

    \caption{
    Overview of \method{}. A shared GNN with learnable resolution embeddings constructs comparable resolution-specific representations and associates each source resolution with a lightweight expert. Cross-Resolution Prototype Transport derives expert-specific routing over target resolutions from cross-domain class-resolution prototypes, while Cross-Resolution Target Grafting mixes class-conditioned target-to-source prototype displacements according to target posteriors.
    }
    \label{fig:framework}

\end{figure*}

\noindent\textbf{Multi-Resolution Graph Representation Learning.} Multi-resolution graph learning captures complementary evidence across different propagation ranges~\cite{yang2022omni,zhang2022graph,wang2025protomol}. Representative methods use multi-hop propagation, layer-wise aggregation, hierarchical pooling, or spectral filtering to model local and higher-order structures~\cite{chen2023nagphormer,wu2026pixdiff,wang2026sgac}. Their resolution-specific features are typically selected, aggregated, or fused into a unified representation to improve predictive quality~\cite{sun2023all,yao2024mugsi,xiang2025jailbreaking}. However, these methods primarily focus on within-domain representation enrichment and do not specify how knowledge acquired at a source propagation resolution should be transferred across target resolutions. Existing GDA methods incorporating multi-resolution information similarly use it to enrich or calibrate representations, rather than learning explicit source-to-target resolution correspondence~\cite{ngo2025higda,shou2025graph}.In contrast, \method{} preserves comparable resolution-specific representations, learns soft cross-resolution correspondence, and applies it through posterior-conditioned target grafting.

%% file: code/4_method.tex
\section{Methodology}

In this paper, we study GDA for graph classification. Let $G=(\mathcal{V},\mathcal{E},\mathbf{X})$ denote a graph, where $\mathcal{V}$ is the node set, $\mathcal{E}\subseteq\mathcal{V}\times\mathcal{V}$ is the edge set, and $\mathbf{X}\in\mathbb{R}^{|\mathcal{V}|\times d_x}$ is the node-feature matrix with $d_x$ features per node. We are given a labeled source domain $\mathcal{D}^{s}=\{(G_i^{s},y_i^{s})\}_{i=1}^{N_s}$ and an unlabeled target domain $\mathcal{D}^{t}=\{G_i^{t}\}_{i=1}^{N_t}$, where $y_i^{s}\in\mathcal{Y}=\{1,\ldots,C\}$. The two domains share the label space $\mathcal{Y}$ but satisfy $\mathbb{P}^{s}(G,Y)\neq\mathbb{P}^{t}(G,Y)$. Let $f_{\Theta}$ denote a graph classifier parameterized by $\Theta$ that produces a distribution over $C$ classes. Its target-domain risk is defined as
\begin{equation}
    \mathcal{R}_{t}(f_{\Theta})
    =
    \mathbb{E}_{(G^{t},Y^{t})\sim\mathbb{P}^{t}}
    \left[
        \ell\!\left(
            f_{\Theta}(G^{t}),
            Y^{t}
        \right)
    \right],
    \label{eq:problem_formulation}
\end{equation}
where $G^{t}$ and $Y^{t}\in\mathcal{Y}$ denote a target graph and its unobserved label, respectively, and $\ell$ is the classification loss. The ideal objective is $\Theta^{\star}\in\arg\min_{\Theta}\mathcal{R}_{t}(f_{\Theta})$.

\subsection{Overview of \method{}}

As shown in Figure~\ref{fig:framework}, we propose \method{}, which learns source-to-target resolution correspondence for GDA. First, \method{} constructs a multi-resolution representation bank with a shared GNN and learnable resolution embeddings, and associates each source resolution with an expert. Second, \textit{Cross-Resolution Prototype Transport} (CRPT) builds class-resolution prototypes from source labels and soft target posteriors, and converts cross-domain prototype discrepancies into expert-specific routing distributions over target resolutions. Third, \textit{Cross-Resolution Target Grafting} (CRTG) converts the global correspondence into target-specific adaptation by mixing class-conditioned target-to-source prototype displacements according to each target posterior.

\subsection{Multi-Resolution Representation Bank and Source Experts}
\label{sec:multi_resolution}

Learning source-to-target resolution correspondence requires representations that preserve resolution-specific evidence while remaining comparable across resolutions and domains~\cite{fan2025towards,fan2025multi}. Accordingly, we construct a multi-resolution representation bank using a shared GNN encoder and learnable resolution embeddings.

For a graph $G=(\mathcal{V},\mathcal{E},\mathbf{X})$, let $n_G=|\mathcal{V}|$ and $\mathbf{A}_G\in\{0,1\}^{n_G\times n_G}$ denote its adjacency matrix. The normalized propagation matrix is defined as
\begin{equation}
    \widetilde{\mathbf{A}}_G=\mathbf{A}_G+\mathbf{I}_{n_G},
    \quad
    \mathbf{S}_G=
    \widetilde{\mathbf{D}}_G^{-\frac{1}{2}}
    \widetilde{\mathbf{A}}_G
    \widetilde{\mathbf{D}}_G^{-\frac{1}{2}},
    \label{eq:normalized_propagation}
\end{equation}
where $\widetilde{\mathbf{D}}_G=\operatorname{diag}(\widetilde{\mathbf{A}}_G\mathbf{1}_{n_G})$ is the degree matrix, $\mathbf{I}_{n_G}$ is the identity matrix, and $\mathbf{1}_{n_G}\in\mathbb{R}^{n_G}$ is the all-ones vector.

Let $\mathcal{J}=\{0,\ldots,J\}$ denote the resolution index set and $\mathcal{R}=(r_0,\ldots,r_J)$ the corresponding nonnegative integer pre-propagation orders, with $0=r_0<\cdots<r_J$. We use $r_j$ to operationalize propagation resolution: $r_0=0$ retains the original node features, while larger $r_j$ aggregate information from broader neighborhoods. The node features at resolution $j$ are
\begin{equation}
    \mathbf{X}_G^{(j)}=\mathbf{S}_G^{\,r_j}\mathbf{X},
    \quad
    j\in\mathcal{J}.
    \label{eq:multi_resolution_features}
\end{equation}

All resolutions are processed by a shared GNN encoder $\Phi_{\theta}$. To preserve resolution identity under parameter sharing, we introduce a learnable resolution-embedding table $\mathbf{E}_{\mathrm{res}}\in\mathbb{R}^{(J+1)\times d_r}$, where $d_r$ is the embedding dimension. Its $j$-th row is $\mathbf{e}_j^{\top}$, with $\mathbf{e}_j\in\mathbb{R}^{d_r}$ identifying pre-propagation order $r_j$. Each resolution embedding is shared across source and target graphs. The graph representation at resolution $j$ is defined as
\begin{equation}
    \mathbf{z}_G^{(j)}
    =
    \operatorname{READOUT}
    \left(
        \Phi_{\theta}
        \left(
            \mathbf{A}_G,\,
            \left[
                \mathbf{X}_G^{(j)}
                \,\Vert\,
                \mathbf{1}_{n_G}\mathbf{e}_j^{\top}
            \right]
        \right)
    \right)
    \in\mathbb{R}^{d},
    \label{eq:multi_resolution_encoding}
\end{equation}
where $[\cdot\Vert\cdot]$ denotes feature concatenation, $\operatorname{READOUT}$ is a permutation-invariant pooling operator~\cite{xu2018how}, and $d$ is the graph-representation dimension. The shared encoder maps all resolution-specific representations into a common latent space, while the resolution embeddings preserve their pre-propagation identities. The resulting multi-resolution representation bank is
\begin{equation}
    \mathcal{Z}_G
    =
    \left(
        \mathbf{z}_G^{(0)},
        \ldots,
        \mathbf{z}_G^{(J)}
    \right).
    \label{eq:representation_bank}
\end{equation}

To capture the discriminative knowledge associated with each source resolution, we assign a source expert $h_j:\mathbb{R}^{d}\rightarrow\mathbb{R}^{C}$ to resolution index $j$:
\begin{equation}
    h_j(\mathbf{z})
    =
    \mathbf{W}_j\mathbf{z}+\mathbf{b}_j,
    \quad
    \mathbf{W}_j\in\mathbb{R}^{C\times d},
    \quad
    \mathbf{b}_j\in\mathbb{R}^{C},
    \label{eq:resolution_expert}
\end{equation}
where $h_j(\mathbf{z})$ is the logit vector over the $C$ classes. During source supervision, $h_j$ is paired with $\mathbf{z}_{G_i^s}^{(j)}$, so index $j$ links pre-propagation order $r_j$, representation $\mathbf{z}_G^{(j)}$, and expert $h_j$.

We use global mixture weights to model the relative contributions of the source experts. Let $\mathbf{a}=[a_0,\ldots,a_J]^{\top}\in\mathbb{R}^{J+1}$ be a learnable score vector. The normalized weight of expert $h_j$ is
\begin{equation}
    \alpha_j
    =
    \frac{\exp(a_j)}
    {\sum_{j'\in\mathcal{J}}\exp(a_{j'})},
    \quad
    j\in\mathcal{J},
    \label{eq:expert_mixture_weights}
\end{equation}
where $\sum_{j\in\mathcal{J}}\alpha_j=1$. Because resolution-specific representations share the same latent space, $h_j(\mathbf{z}_{G_i^t}^{(k)})$ is well-defined for any $k\in\mathcal{J}$, enabling cross-resolution routing.

\subsection{Cross-Resolution Prototype Transport}
\label{sec:crpt}

The multi-resolution representation bank provides resolution-specific representations, but source-to-target correspondence remains unknown. To address this issue, we introduce Cross-Resolution Prototype Transport (CRPT), which iteratively estimates soft target class structure and refines prototype-guided routing.

To encode resolution correspondence and aggregate cross-resolution expert responses, we introduce a row-stochastic matrix $\boldsymbol{\Gamma}\in\mathbb{R}_{+}^{(J+1)\times(J+1)}$. The entry $\Gamma_{j,k}$ weights the response of source expert $h_j$ when evaluated on target resolution $k$, subject to
\begin{equation}
    \Gamma_{j,k}\geq 0,
    \quad
    \forall j,k\in\mathcal{J},
    \quad
    \sum_{k\in\mathcal{J}}\Gamma_{j,k}=1,
    \quad
    \forall j\in\mathcal{J}.
    \label{eq:correspondence_constraint}
\end{equation}
The $j$-th row of $\boldsymbol{\Gamma}$ defines a soft routing distribution over target resolutions for expert $h_j$. We initialize it uniformly as $\Gamma_{j,k}=1/(J+1)$.

Using the current routing matrix, we infer soft class posteriors for target graphs by aggregating expert responses across target resolutions:
\begin{equation}
    \begin{aligned}
        \boldsymbol{\ell}_i^t
        =
        \sum_{j\in\mathcal{J}}
        \sum_{k\in\mathcal{J}}
        \alpha_j\Gamma_{j,k}
        h_j\!\left(
            \mathbf{z}_{G_i^t}^{(k)}
        \right),\quad
        \mathbf{p}_i^t
        =
        \sigma\!\left(
            \boldsymbol{\ell}_i^t
        \right),
    \end{aligned}
    \label{eq:target_prediction}
\end{equation}
where $\sigma(\cdot)$ denotes the softmax function and $\mathbf{p}_i^t=[p_{i,1}^t,\ldots,p_{i,C}^t]^\top$ is the predicted class distribution of target graph $G_i^t$. These posteriors provide soft memberships for estimating class structure at each target resolution. Together with source labels, they define class-resolution prototypes. Let $N_{s,c}=|\{i\mid y_i^s=c\}|$ denote the number of source samples in class $c$. For each $c\in\mathcal{Y}$ and $j,k\in\mathcal{J}$, we define
\begin{equation}
    \begin{aligned}
        \boldsymbol{\mu}_{c,j}^{s}
        =
        \frac{1}{N_{s,c}}
        \sum_{i:y_i^s=c}
        \mathbf{z}_{G_i^s}^{(j)},\quad
        \boldsymbol{\mu}_{c,k}^{t}
        =
        \frac{
            \sum_{i=1}^{N_t}
            p_{i,c}^{t}\,
            \mathbf{z}_{G_i^t}^{(k)}
        }{
            \sum_{i=1}^{N_t}
            p_{i,c}^{t}
            +\varepsilon
        },
    \end{aligned}
    \label{eq:class_resolution_prototypes}
\end{equation}
where $\varepsilon>0$ ensures numerical stability. Here, $\boldsymbol{\mu}_{c,j}^{s}$ is the source class center at resolution $j$, while $\boldsymbol{\mu}_{c,k}^{t}$ is its probability-weighted target counterpart at resolution $k$.

As $h_j$ exclusively processes source representations at resolution $j$ in mixture-based training, target resolutions with similar class-conditional geometry are more likely to support its resolution-indexed decision function. We therefore measure the compatibility between source resolution $j$ and target resolution $k$ using the class-averaged prototype discrepancy
\begin{equation}
    D_{j,k}
    =
    \frac{1}{C}
    \sum_{c=1}^{C}
    \left\|
        \boldsymbol{\mu}_{c,j}^{s}
        -
        \boldsymbol{\mu}_{c,k}^{t}
    \right\|_2^2,
    \label{eq:resolution_discrepancy}
\end{equation}
where a smaller $D_{j,k}$ indicates closer class-conditional structure. We convert the discrepancy matrix into a refined routing matrix through row-wise softmax normalization:
\begin{equation}
    \Gamma_{j,k}^{+}
    =
    \frac{
        \exp\!\left(-D_{j,k}/\tau\right)
    }{
        \sum_{k'\in\mathcal{J}}
        \exp\!\left(-D_{j,k'}/\tau\right)
    },
    \quad
    j,k\in\mathcal{J},
    \label{eq:resolution_correspondence}
\end{equation}
where $\tau>0$ controls the routing concentration. A larger $\Gamma_{j,k}^{+}$ gives expert $h_j$ greater weight on target resolution $k$. Since $D_{j,k}$ is averaged over classes, the resulting routing is informed by class structure but shared across classes.

Finally, we define a routing-weighted prototype-matching objective to reduce cross-domain discrepancies under the refined correspondence:
\begin{equation}
    \mathcal{L}_{\mathrm{CRPT}}
    =
    \frac{1}{J+1}
    \sum_{j\in\mathcal{J}}
    \sum_{k\in\mathcal{J}}
    \Gamma_{j,k}^{+}D_{j,k}.
    \label{eq:crpt_loss}
\end{equation}
Within each adaptation step, the current $\boldsymbol{\Gamma}$ is used to infer target posteriors, after which $\boldsymbol{\Gamma}^{+}$ is computed from the resulting prototypes, weights the adaptation objectives, and becomes the routing matrix for the next step.

\subsection{Cross-Resolution Target Grafting}
\label{sec:crtg}

Class-resolution prototypes capture cross-domain class structure, while the refined routing matrix encodes global compatibility between source and target resolutions. However, this class-shared correspondence does not specify how to adapt an individual target graph under uncertain class membership. We therefore introduce Cross-Resolution Target Grafting (CRTG), which translates global correspondence into instance-specific adaptation through posterior-conditioned prototype displacements.

For each class $c\in\mathcal{Y}$ and source--target resolution pair $(j,k)$, we define the target-to-source prototype displacement as
\begin{equation}
    \boldsymbol{\Delta}_{c}^{j\leftarrow k}
    =
    \boldsymbol{\mu}_{c,j}^{s}
    -
    \boldsymbol{\mu}_{c,k}^{t}.
    \label{eq:class_prototype_displacement}
\end{equation}
Because source and target representations share the same latent space, this displacement is well-defined. The notation $j\leftarrow k$ indicates a translation from the target class prototype at resolution $k$ toward the corresponding source prototype at resolution $j$.

For target graph $G_i^t$, we combine the class-conditioned displacements using its current posterior:
\begin{equation}
    \boldsymbol{\Delta}_{i}^{j\leftarrow k}
    =
    \sum_{c=1}^{C}
    p_{i,c}^{t}
    \boldsymbol{\Delta}_{c}^{j\leftarrow k},
    \label{eq:sample_prototype_displacement}
\end{equation}
where $p_{i,c}^{t}$ is the probability assigned to class $c$. The resulting $\boldsymbol{\Delta}_{i}^{j\leftarrow k}\in\mathbb{R}^{d}$ provides a posterior-weighted transfer direction that preserves class uncertainty without requiring a hard assignment. We then graft the target representation at resolution $k$ toward the source class structure at resolution $j$:
\begin{equation}
    \begin{aligned}
        \widetilde{\mathbf{z}}_{G_i^t}^{j\leftarrow k}
        =
        \mathbf{z}_{G_i^t}^{(k)}
        +
        \beta
        \boldsymbol{\Delta}_{i}^{j\leftarrow k},\quad
        \mathbf{q}_{i}^{j\leftarrow k}
        =
        \sigma\!\left(
            h_j\!\left(
                \widetilde{\mathbf{z}}_{G_i^t}^{j\leftarrow k}
            \right)
        \right),
    \end{aligned}
    \label{eq:grafted_prediction}
\end{equation}
where $\beta\in[0,1]$ controls the grafting magnitude and $\mathbf{q}_{i}^{j\leftarrow k}$ is the class distribution predicted by expert $h_j$ from the grafted representation.

Finally, we define a routing-weighted prediction-consistency objective that encourages grafted predictions from compatible resolution pairs to agree with the aggregate target posterior:
\begin{equation}
    \mathcal{L}_{\mathrm{CRTG}}
    =
    \frac{1}{N_t(J+1)}
    \sum_{i=1}^{N_t}
    \sum_{j\in\mathcal{J}}
    \sum_{k\in\mathcal{J}}
    \Gamma_{j,k}^{+}
    \operatorname{KL}\!\left(
        \mathbf{p}_i^t
        \,\middle\|\,
        \mathbf{q}_{i}^{j\leftarrow k}
    \right),
    \label{eq:crtg_loss}
\end{equation}
where $\operatorname{KL}(\cdot\|\cdot)$ denotes the Kullback--Leibler divergence.

\subsection{Learning Objective}
\label{sec:objective}

Source supervision trains the source-expert mixture on the representations associated with their respective resolutions. For source graph $G_i^s$, the class-logit vector and predicted distribution are
\begin{equation}
    \begin{aligned}
        \boldsymbol{\ell}_i^s
        =
        \sum_{j\in\mathcal{J}}
        \alpha_j
        h_j\!\left(
            \mathbf{z}_{G_i^s}^{(j)}
        \right),\quad
        \mathbf{p}_i^s
        =
        \sigma\!\left(
            \boldsymbol{\ell}_i^s
        \right),
    \end{aligned}
    \label{eq:source_prediction}
\end{equation}
where $\mathbf{p}_i^s$ is the predicted source class distribution. The supervised cross-entropy loss is
\begin{equation}
    \mathcal{L}_{\mathrm{sup}}
    =
    -\frac{1}{N_s}
    \sum_{i=1}^{N_s}
    \log p_{i,y_i^s}^{s},
    \label{eq:supervised_loss}
\end{equation}
where $p_{i,y_i^s}^{s}$ is the probability assigned to the ground-truth label $y_i^s$. 

The overall objective combines source supervision, prototype-guided cross-resolution correspondence learning, and posterior-conditioned target grafting:
\begin{equation}
    \mathcal{L}
    =
    \mathcal{L}_{\mathrm{sup}}
    +
    \lambda_{CRPT}\mathcal{L}_{\mathrm{CRPT}}
    +
    \lambda_{CRTG}\mathcal{L}_{\mathrm{CRTG}},
    \label{eq:overall_objective}
\end{equation}
where $\lambda_{CRPT}\geq 0$ and $\lambda_{CRTG}\geq 0$ balance the CRPT and CRTG objectives, respectively. 

\subsection{Theoretical Analysis}

\begin{theorem}[Target-Risk Decomposition for \method{}]
\label{thm:generalization}
Let \(\mathfrak T\) be a class of complete \method{} states fixed independently of the labeled source sample. Define
\(f_\Theta^s(G):=\sigma\!\left(\sum_{j\in\mathcal J}\alpha_jh_j(\mathbf z_G^{(j)})\right)\) and
\(f_\Theta(G):=\sigma\!\left(\sum_{j,k\in\mathcal J}\alpha_j\Gamma_{j,k}h_j(\mathbf z_G^{(k)})\right)\).
Assume uniformly over states, experts, and classes that logits on original and grafted representations are bounded, \(\beta\in[0,1]\), \(\pi_s(c),\pi_t(c)>0\), \(0\leq\ell\leq B_\ell\), \(\ell(\cdot,y)\) is \(L_p\)-Lipschitz in \(\ell_1\), \(z\mapsto\ell(\sigma(h_j(z)),y)\) is \(L_z\)-Lipschitz in \(\ell_2\), and class-conditional representations have finite first moments. For every \(\delta\in(0,1)\), with probability at least \(1-\delta\) over \(S_s\sim(\mathbb P^s)^{N_s}\), simultaneously for all \(\Theta\in\mathfrak T\),
\begin{equation}
\label{eq:generalization_bound}
\begin{aligned}
\mathcal R_t(f_\Theta)
&\leq{}\widehat{\mathcal R}_s(f_\Theta^s)
+2\widehat{\mathfrak R}_{S_s}(\ell\circ\mathcal F_s)
+3B_\ell\sqrt{\frac{\log(2/\delta)}{2N_s}}\\
&+K_{\rm res}\sqrt{\mathfrak D_{\rm CRPT}^{\rm pop}(\Theta)}
+K_{\rm graft}\sqrt{\mathfrak C_{\rm CRTG}^{\rm pop}(\Theta)}
+\Lambda_\Theta,
\end{aligned}
\end{equation}
where
\(\widehat{\mathcal R}_s(f):=N_s^{-1}\sum_{i=1}^{N_s}\ell(f(G_i^s),y_i^s)\),
\(\mathcal F_s:=\{f_\Theta^s:\Theta\in\mathfrak T\}\),
\(\pi_d(c):=\mathbb P^d(Y^d=c)\) for \(d\in\{s,t\}\),
\(\pi_{\max}^t:=\max_c\pi_t(c)\), and
\(\alpha_\star:=\sup_{\Theta\in\mathfrak T}\max_{j\in\mathcal J}\alpha_j\leq1\).
The constants are
\(K_{\rm res}:=(1-\beta)L_z\sqrt{C\pi_{\max}^t\alpha_\star(J+1)}\)
and
\(K_{\rm graft}:=L_p\sqrt{2\alpha_\star(J+1)}\).
Moreover, \(\mathfrak D_{\rm CRPT}^{\rm pop}(\Theta)\) and
\(\mathfrak C_{\rm CRTG}^{\rm pop}(\Theta)\) denote the population-level
prototype discrepancy and grafting consistency under the refined routing,
while $\Lambda_\Theta$ collects residual prior, conditional-shape,
posterior/prototype, and expert-mixture errors.
\end{theorem}

The theorem shows that the target risk is controlled by the source risk, cross-resolution prototype discrepancy, and grafting inconsistency, implying that minimizing these terms tightens the bound when the residual terms are sufficiently small.

%% file: code/5_exp.tex

\section{Experiments}

\subsection{Experimental Settings}

\noindent\textbf{Datasets.} To evaluate the effectiveness of \method{}, we conduct experiments on graph classification benchmarks under two representative types of domain shifts. (1) \textbf{Structure-based shifts:} We use Mutagenicity~\cite{kazius2005derivation}, PROTEINS~\cite{dobson2003distinguishing}, NCI1~\cite{wale2008comparison}, and ogbg-molhiv~\citep{hu2021ogblsc}. Following~\cite{yin2022deal,yin2025coupling}, each dataset is partitioned into multiple subdomains according to node and edge densities. (2) \textbf{Feature-based shifts:} We further evaluate on PROTEINS, DD, BZR, BZR\_MD, COX2, and COX2\_MD~\cite{sutherland2003spline}, where the source and target domains primarily differ in their node feature distributions. 

\input{table/main_results}

\noindent\textbf{Baselines.}
We compare \method{} against representative baselines spanning three categories: (1) graph kernels and path-based models, including the WL subtree kernel~\cite{shervashidze2011weisfeiler} and PathNN~\cite{michel2023path}; (2) general-purpose graph neural networks, including GCN~\cite{kipf2016semi}, GIN~\cite{xu2018how}, GMT~\cite{baek2021accurate}, and CIN~\cite{bodnar2021weisfeiler}; and (3) graph domain adaptation methods, including DEAL~\cite{yin2022deal}, SGDA~\cite{qiao2023semi}, A2GNN~\cite{liu2024rethinking}, StruRW~\cite{liu2023structural}, PA-BOTH~\cite{liu2024pairwise}, GAA~\cite{fang2025benefits}, and TDSS~\cite{chen2025smoothness}. 

\noindent\textbf{Implementation Details.}
We implement \method{} in PyTorch on four NVIDIA GeForce RTX 3090 GPUs. The shared encoder $\Phi_{\theta}$ is a three-layer GIN~\cite{xu2018how}, using representation dimension $d=128$, resolution-embedding dimension $d_r=16$, and dropout 0.2. The multi-resolution representation bank uses $J=3$ and $\mathcal{R}=(r_0,\ldots,r_J)=(0,1,2,4)$. We optimize the model using Adam with batch size 64 and learning rate $5\times10^{-4}$. We set the CRPT and CRTG weights to $\lambda_{\text{CRPT}}=10^{-2}$ and $\lambda_{\text{CRTG}}=10^{-3}$, respectively, and the grafting strength to $\beta=0.5$. Target labels are not used for hyperparameter tuning and only for post-hoc evaluation. All results are averaged over five independent runs for fair comparisons.

\begin{figure}[t]
    \centering
    \begin{subfigure}[t]{0.23\textwidth}
        \centering
        \includegraphics[width=\linewidth]{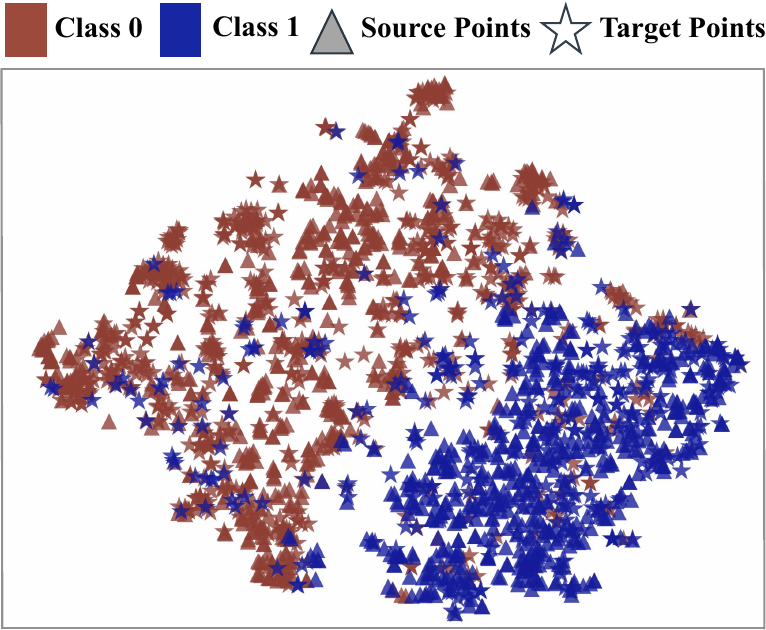}
        \caption{\method{}}
        \label{fig:ours}
    \end{subfigure}\hfill
    \begin{subfigure}[t]{0.23\textwidth}
        \centering
    \raisebox{0.0cm}{\includegraphics[width=\linewidth]{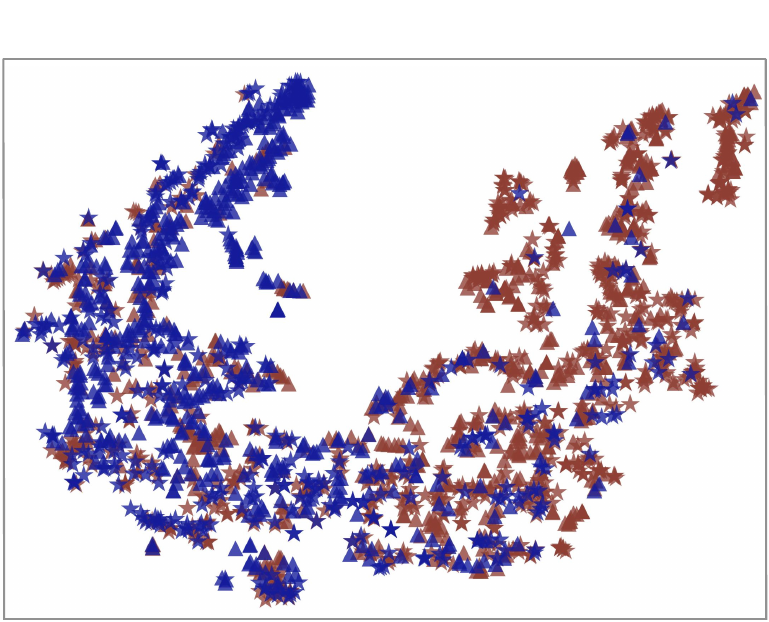}}
        \caption{GAA}\label{fig:tsne_gaa}
    \end{subfigure}
    \vspace{-0.2cm}
    \caption{T-SNE visualizations of \method{} and the baseline GAA on the Mutagenicity dataset.}
    \vspace{-0.3cm}
    \label{fig:generalization_tsne_mutag}
\end{figure}

\subsection{Performance Comparison}

We report the performance of \method{} and all baselines under different domain shifts in Tables~\ref{tab:main_results} and \ref{tab:proteins_node_shift}--\ref{tab:molhiv_edge_shift}. The results support the following observations. (1) General-purpose GNNs outperform the WL subtree kernel, particularly under structural shifts, suggesting that learned graph representations transfer more effectively than handcrafted kernel similarities in these settings. Nevertheless, their direct transfer performance remains limited because they do not explicitly account for cross-domain distribution gaps or semantic resolution shift. (2) GDA methods generally improve over non-adaptive models through representation alignment, structural reweighting, or propagation regularization. However, their gains vary across shifts, suggesting that representation-level alignment or domain-level propagation calibration may be insufficient when class-discriminative evidence moves across resolutions. (3) \method{} achieves the best performance on most tasks under both structural and feature shifts. These gains are consistent with its design: comparable multi-resolution representations preserve resolution-specific evidence, CRPT learns class-structure-informed soft correspondence across resolutions, and CRTG converts the global correspondence into posterior-conditioned target adaptation. Together, these components enable source experts to exploit compatible target resolutions rather than fixed same-resolution pairs. In addition, Figure~\ref{fig:generalization_tsne_mutag} compares the t-SNE visualizations of \method{} and GAA. \method{} produces more compact and better-separated target class clusters, providing qualitative support for improved target discriminability.

\begin{figure}[t]
    \centering

    \begin{subfigure}[t]{0.48\columnwidth}
        \centering
        \includegraphics[width=\linewidth]{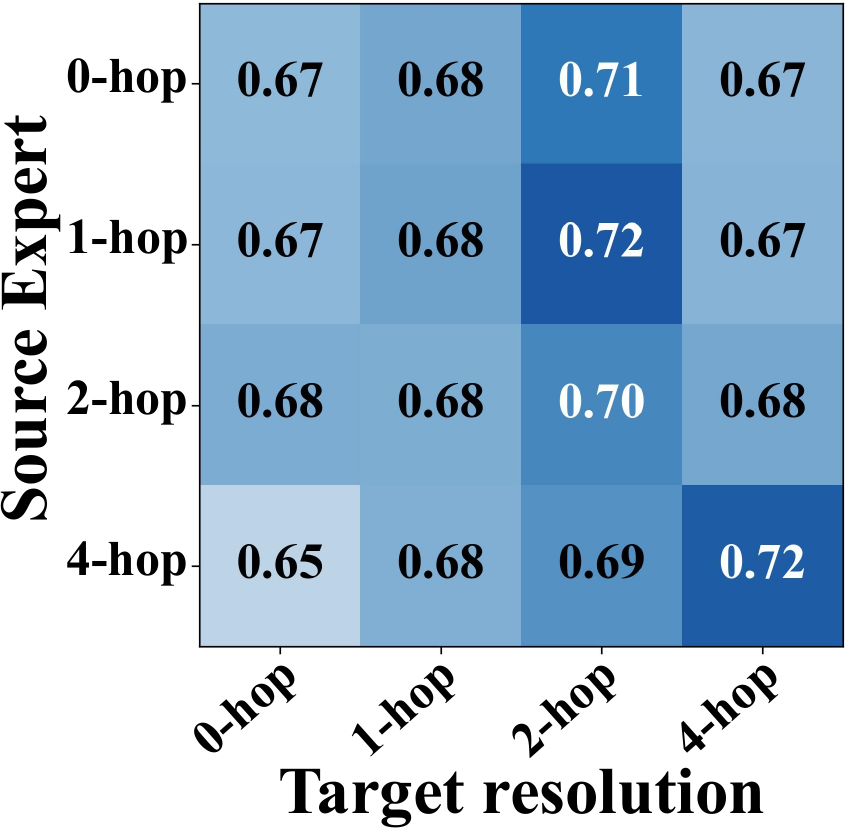}
        \caption{Oracle compatibility $\mathbf{O}$}
        \label{fig:oracle_compatibility}
    \end{subfigure}
    \hfill
    \begin{subfigure}[t]{0.48\columnwidth}
        \centering
        \includegraphics[width=\linewidth]{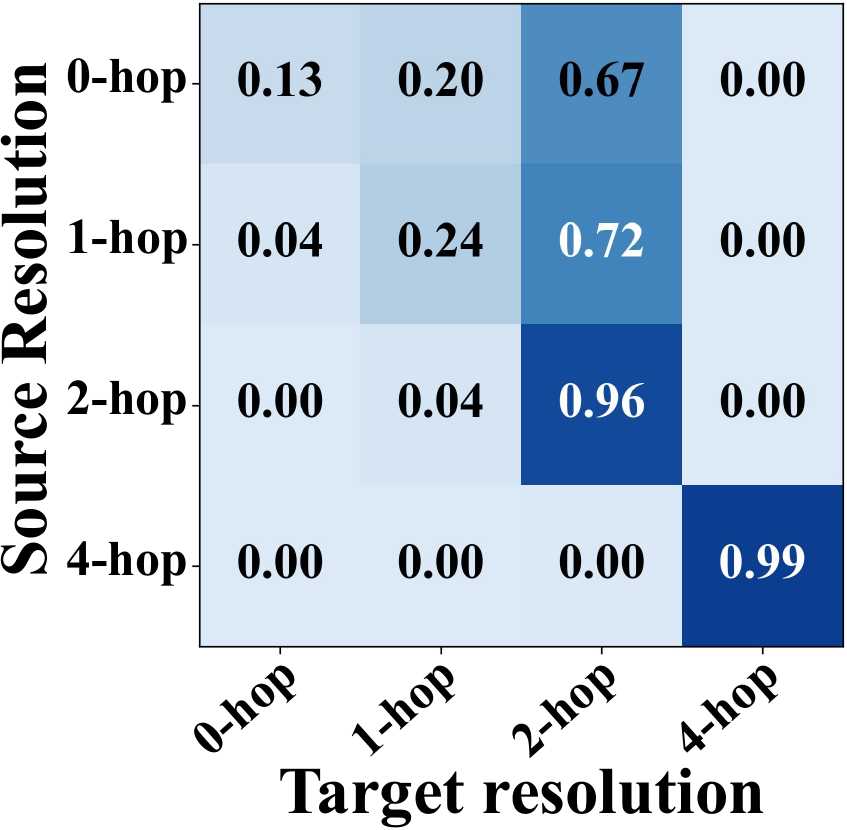}
        \caption{Learned correspondence $\boldsymbol{\Gamma}$}
        \label{fig:learned_correspondence}
    \end{subfigure}
    \vspace{-0.2cm}
    \caption{Oracle compatibility and learned cross-resolution correspondence matrices for \method{} on the Mutagenicity dataset.}
    \label{fig:oracle_correspondence_validation}
    \vspace{-0.3cm}
\end{figure}

\subsection{Cross-Resolution Correspondence Analysis}

To assess whether the learned correspondence captures label-informed expert--resolution compatibility, we construct a post-hoc oracle matrix $\mathbf{O}$, where $O_{j,k}$ denotes the target accuracy of source expert $h_j$ evaluated on target representations at resolution $k$. We compare its row-wise preferences with the correspondence $\boldsymbol{\Gamma}$ learned without target labels. As shown in Figure~\ref{fig:oracle_correspondence_validation}, both matrices select the 2-hop, 2-hop, 2-hop, and 4-hop target resolutions for the 0-hop, 1-hop, 2-hop, and 4-hop source experts, respectively, yielding a $4/4$ ($100\%$) Top-1 agreement. The 0-hop and 1-hop experts prefer the 2-hop target representation, indicating cross-resolution shifts, whereas the 2-hop and 4-hop experts retain same-resolution matches. Thus, on this task, \method{} matches the oracle-preferred target resolution for all source experts, with target labels used only for post-hoc evaluation.

\begin{figure}[t]
    \centering

    \begin{subfigure}[b]{0.48\columnwidth}
    \centering
    \includegraphics[
        height=3.2cm,
        keepaspectratio
    ]{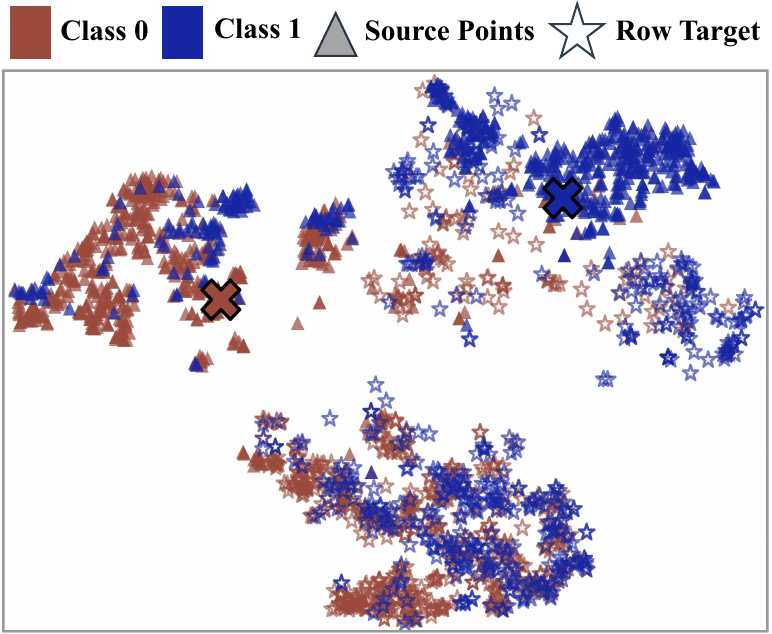}
    \caption{Source and raw target representations before grafting.}
    \label{fig:graft_a}
\end{subfigure}
\hfill
\begin{subfigure}[b]{0.48\columnwidth}
    \centering
    \includegraphics[
        height=3.2cm,
        keepaspectratio
    ]{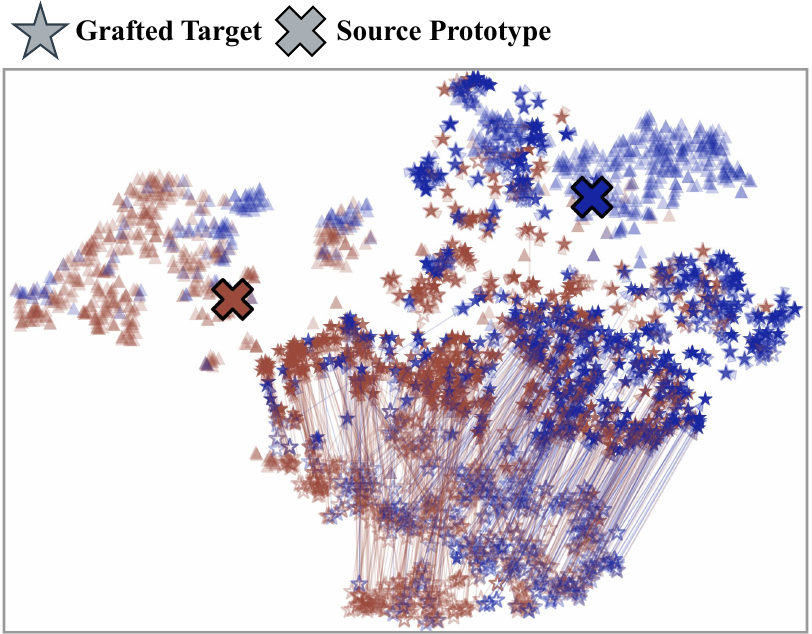}
    \caption{Posterior-conditioned target displacements.}
    \label{fig:graft_b}
\end{subfigure}

    \caption{Visualization of cross-resolution target grafting on Mutagenicity.}
    \label{fig:grafting}
    \vspace{-0.3cm}
\end{figure}

\subsection{Cross-Resolution Target Grafting Analysis}

To examine how the learned global correspondence guides individual target graphs, we visualize target representations before and after grafting on the Mutagenicity dataset. Figure~\ref{fig:grafting}(a) shows that raw target representations remain dispersed relative to the source class regions, indicating residual instance-level mismatch before CRTG. Figure~\ref{fig:grafting}(b) shows that grafting induces sample-dependent movements toward different source regions rather than a uniform translation. These movements arise from posterior-weighted combinations of class-conditioned target-to-source prototype displacements. Thus, CRTG translates the global, class-structure-informed correspondence into instance-specific representation adjustments under class uncertainty, providing qualitative support for the intended global-to-instance transfer mechanism.

\begin{table}[h]
\centering
\resizebox{0.48\textwidth}{!}{
\begin{tabular}{l|c|c|c|c|c|c}
\toprule
\textbf{Methods}
& M0$\rightarrow$M1 & M1$\rightarrow$M0
& M0$\rightarrow$M2 & M2$\rightarrow$M0
& M0$\rightarrow$M3 & M3$\rightarrow$M0 \\
\midrule 
\method{} w/o MR
& 72.9 & 71.1 & 58.7 & 71.3 & 54.5 & 52.9 \\
\method{} w/o CRR
& 73.2 & 71.9 & 59.6 & 72.9 & 54.5 & 53.4 \\
\method{} w/o CRPT
& 75.7 & 72.2 & 60.3 & 71.3 & 53.6 & 53.9 \\
\method{} w/o CRTG
& 75.9 & 74.5 & 68.5 & 72.9 & 63.2 & 57.9 \\
\method{} w/ HG
& 78.4 & 75.4 & 71.4 & 73.5 & 66.2 & 60.7 \\
\midrule
\method{}
& \textbf{80.7}
& \textbf{76.9}
& \textbf{72.9}
& \textbf{74.5}
& \textbf{67.4}
& \textbf{63.4} \\
\bottomrule
\end{tabular}
}
\caption{The results of ablation studies on the Mutagenicity dataset. \textbf{Bold} results indicate the best performance.} 
\label{tab:ablation}
\vspace{-0.2cm}
\end{table}

\subsection{Ablation Study}

To assess the contribution of each component in \method{}, we evaluate five variants: (1) \method{} w/o MR replaces the multi-resolution representation bank with a single-resolution representation; (2) \method{} w/o CRR replaces the learned cross-resolution routing with fixed same-resolution pairing; (3) \method{} w/o CRPT retains cross-resolution routing but removes the prototype-matching objective; (4) \method{} w/o CRTG removes posterior-conditioned target grafting; and (5) \method{} w/ HG replaces posterior-weighted displacement mixing with hard grafting, which selects the displacement of the maximum-posterior class. As shown in Table~\ref{tab:ablation}, we make the following observations. (1) Removing MR or CRR consistently degrades performance, demonstrating that complementary resolution-specific evidence and learned cross-resolution routing are both essential, while fixed same-resolution pairing cannot adequately capture shifted resolution compatibility. (2) Removing CRPT or CRTG also leads to clear performance drops, confirming their complementary roles in aligning cross-domain class structure and resolving residual instance-level target mismatch, respectively. (3) HG outperforms the variant without CRTG but remains inferior to the complete model, showing that grafting itself is beneficial, while posterior-weighted displacement mixing provides more reliable adaptation than committing each target graph to a single pseudo-class.

\begin{figure}[t]
    \centering
    \begin{subfigure}[t]{0.48\columnwidth}
        \centering
        \includegraphics[width=\linewidth]
        {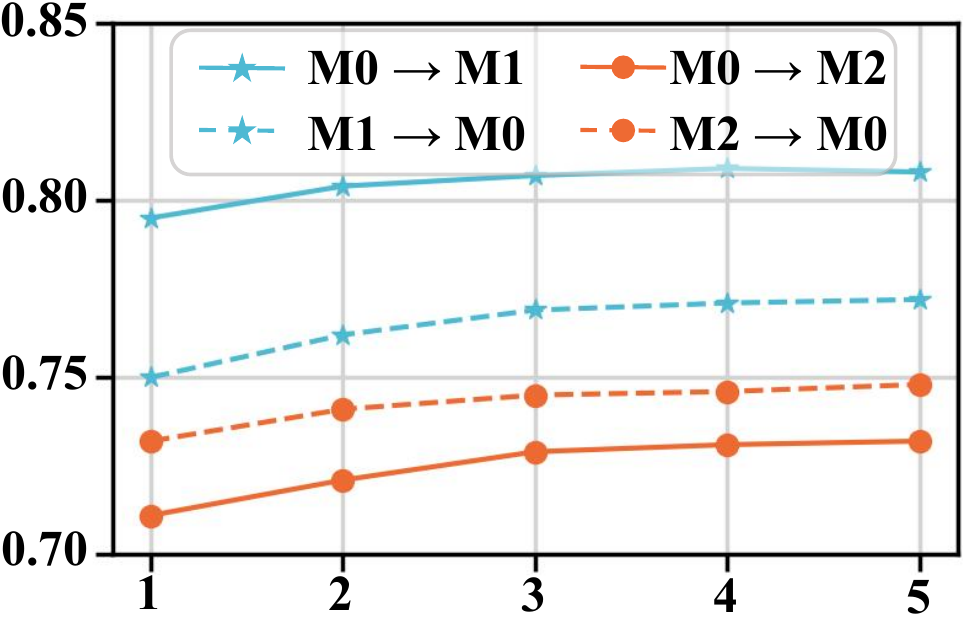}
        \caption{Resolution depth $J$}
        \label{fig:mutagenicity_depth}
    \end{subfigure}
    \hfill
    \begin{subfigure}[t]{0.48\columnwidth}
        \centering
        \includegraphics[width=\linewidth]
        {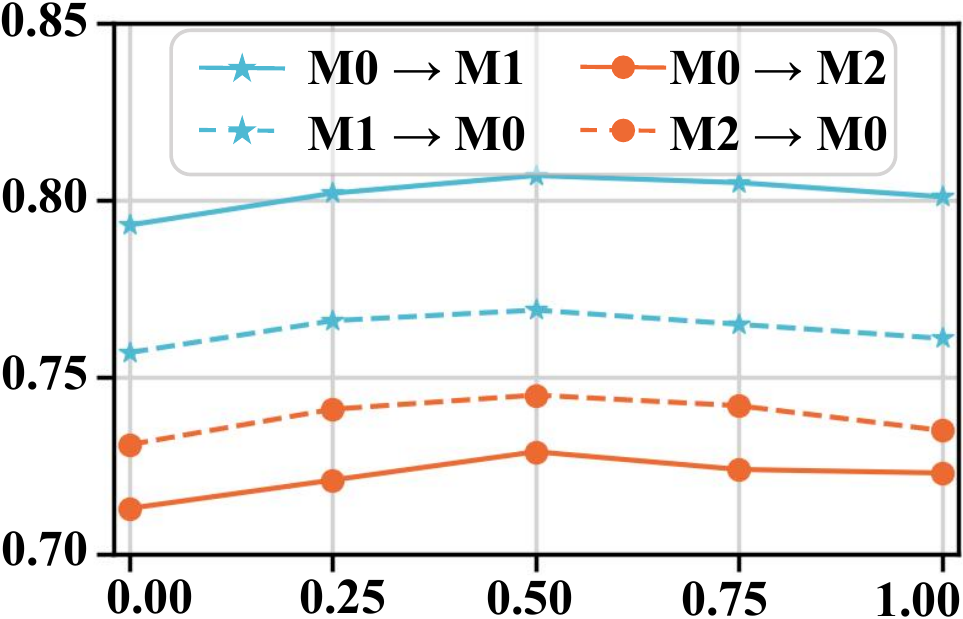}
        \caption{Grafting strength $\beta$}
        \label{fig:mutagenicity_beta}
    \end{subfigure}

    \caption{
        Sensitivity of \method{} to resolution depth $J$ and
        grafting strength $\beta$ on the Mutagenicity dataset.
    }
    \label{fig:sensitivity}
    \vspace{-0.3cm}
\end{figure}

\subsection{Sensitivity Analysis}

We analyze the sensitivity of \method{} to two key hyperparameters: the resolution parameter $J$ and the grafting strength $\beta$. As shown in Figure~\ref{fig:sensitivity}(a), performance improves as $J$ increases and reaches its best level at $J=3$. A small $J$ provides insufficient neighborhood ranges for identifying cross-resolution correspondence, whereas further increasing $J$ introduces limited complementary information with additional computational cost. We then fix $J=3$ and vary $\beta$. As shown in Figure~\ref{fig:sensitivity}(b), performance initially improves and subsequently declines, with the best results obtained at $\beta=0.5$. This trend reflects a trade-off in target grafting: insufficient adjustment leaves sample-level mismatch unresolved, while excessive adjustment may distort the original target semantics and amplify estimation errors. Accordingly, we adopt $J=3$ and $\beta=0.5$ by default, which are pre-specified and
fixed across all tasks.

%% file: table/main_results.tex

\begin{table*}[t]
\centering
\scriptsize
\resizebox{\textwidth}{!}{
\setlength{\tabcolsep}{1.2pt}
\begin{tabular}{
>{\centering\arraybackslash}m{1.8cm}|
*{3}{>{\centering\arraybackslash}m{1.25cm}}|
*{3}{>{\centering\arraybackslash}m{1.25cm}}|
*{6}{>{\centering\arraybackslash}m{1.25cm}}
}
\toprule
\multirow{2}{*}[-0.25em]{\textbf{Methods}}
& \multicolumn{3}{c|}{\textbf{Node Shift}}
& \multicolumn{3}{c|}{\textbf{Edge Shift}}
& \multicolumn{6}{c}{\textbf{Feature Shift}} \\
\cmidrule(lr){2-4} \cmidrule(lr){5-7} \cmidrule(lr){8-13}
& M0$\rightarrow$M1 & M0$\rightarrow$M2 & M0$\rightarrow$M3
& M0$\rightarrow$M1 & M0$\rightarrow$M2 & M0$\rightarrow$M3
& P$\rightarrow$D & D$\rightarrow$P & C$\rightarrow$CM & CM$\rightarrow$C & B$\rightarrow$BM & BM$\rightarrow$B \\
\midrule
WL subtree & 34.3 & 40.4 & 52.7 & 34.4 & 47.6 & 52.7 & 43.0 & 42.2 & 53.1 & 58.2 & 51.3 & 44.0 \\
GCN & 64.1$\pm$1.4 & 65.5$\pm$2.0 & 56.9$\pm$2.1 & 66.3$\pm$1.7 & 63.6$\pm$1.4 & 56.0$\pm$1.4 & 48.9$\pm$2.0 & 60.9$\pm$2.3 & 51.2$\pm$1.8 & 66.9$\pm$1.8 & 48.7$\pm$2.0 & 78.8$\pm$1.7 \\
GIN & 66.5$\pm$2.1 & 52.0$\pm$1.7 & 53.7$\pm$1.7 & 67.1$\pm$1.7 & 54.2$\pm$2.6 & 55.4$\pm$1.9 & 57.3$\pm$2.2 & 61.9$\pm$1.9 & 53.8$\pm$2.5 & 55.6$\pm$2.0 & 49.9$\pm$2.4 & 79.2$\pm$2.8 \\
GMT & 65.7$\pm$1.8 & 62.1$\pm$2.1 & 59.0$\pm$2.0 & 67.9$\pm$1.3 & 61.5$\pm$1.8 & 58.2$\pm$2.4 & 59.5$\pm$2.5 & 50.7$\pm$2.2 & 49.3$\pm$1.8 & 58.2$\pm$2.0 & 50.2$\pm$2.3 & 74.4$\pm$1.8 \\
CIN & 65.1$\pm$1.7 & 66.0$\pm$1.7 & 55.2$\pm$1.5 & 66.3$\pm$1.8 & 60.8$\pm$1.7 & 55.8$\pm$2.4 & 59.1$\pm$2.6 & 58.0$\pm$2.7 & 51.2$\pm$2.0 & 55.6$\pm$1.5 & 49.2$\pm$1.4 & 74.2$\pm$1.9 \\
PathNN & 70.2$\pm$1.5 & 67.1$\pm$2.0 & 58.0$\pm$1.9 & 68.9$\pm$1.9 & 62.9$\pm$1.7 & 58.1$\pm$1.6 & 57.9$\pm$1.8 & 53.8$\pm$3.3 & 49.8$\pm$1.7 & 66.9$\pm$2.5 & 50.3$\pm$2.3 & 75.3$\pm$2.2 \\
\midrule
DEAL & 77.1$\pm$0.9 & 70.9$\pm$0.9 & 60.3$\pm$1.1 & 76.6$\pm$1.6 & 70.6$\pm$1.2 & 60.2$\pm$2.1 & 61.7$\pm$2.0 & 60.0$\pm$1.5 & 52.7$\pm$2.7 & 69.4$\pm$2.9 & 52.4$\pm$2.9 & 78.6$\pm$1.4 \\
SGDA & 77.5$\pm$0.6 & 69.7$\pm$0.5 & 65.5$\pm$0.8 & 75.9$\pm$1.6 & 68.9$\pm$0.8 & 64.4$\pm$0.4 & 48.3$\pm$2.0 & 55.8$\pm$2.6 & 49.8$\pm$1.8 & 66.9$\pm$2.3 & 50.3$\pm$2.1 & 78.8$\pm$2.6 \\
A2GNN & 73.5$\pm$1.9 & 66.1$\pm$1.5 & 60.4$\pm$1.1 & 69.5$\pm$1.4 & 68.6$\pm$1.4 & 58.8$\pm$2.2 & 57.8$\pm$2.1 & 60.3$\pm$1.5 & 51.5$\pm$1.8 & 67.7$\pm$2.1 & 51.6$\pm$2.3 & 77.5$\pm$1.9 \\
StruRW & 78.3$\pm$1.3 & 69.7$\pm$1.3 & 62.6$\pm$0.7 & 76.1$\pm$1.5 & 69.0$\pm$1.3 & 62.1$\pm$1.0 & 59.1$\pm$2.3 & 58.8$\pm$2.8 & 51.2$\pm$2.0 & 54.8$\pm$2.9 & 49.2$\pm$1.4 & 74.7$\pm$2.1 \\
PA-BOTH & 69.8$\pm$1.5 & 63.8$\pm$1.9 & 55.3$\pm$1.1 & 74.7$\pm$1.1 & 65.3$\pm$1.3 & 52.2$\pm$1.5 & 54.2$\pm$3.2 & 56.7$\pm$2.6 & 52.9$\pm$2.8 & 61.8$\pm$2.0 & 47.5$\pm$3.0 & 78.8$\pm$1.9 \\
GAA & 79.3$\pm$1.2 & 71.2$\pm$0.7 & 65.6$\pm$1.3 & 77.5$\pm$1.2 & 70.0$\pm$1.2 & 66.5$\pm$1.3 & 62.4$\pm$0.6 & 64.1$\pm$0.8 & 59.4$\pm$1.8 & 78.4$\pm$1.2 & 57.2$\pm$3.3 & 78.5$\pm$3.0 \\
TDSS & 63.6$\pm$1.3 & 56.7$\pm$1.6 & 54.7$\pm$1.0 & 71.6$\pm$1.5 & 67.3$\pm$1.0 & 55.4$\pm$1.6 & 61.9$\pm$1.1 & 63.6$\pm$1.9 & 56.8$\pm$1.3 & 77.0$\pm$2.6 & 56.6$\pm$1.1 & 79.1$\pm$2.4 \\
\midrule
Ours & \textbf{83.0$\pm$0.8} & \textbf{74.6$\pm$1.3} & \textbf{66.2$\pm$0.5} & \textbf{80.7$\pm$0.9} & \textbf{72.9$\pm$1.7} & \textbf{67.4$\pm$1.3} & \textbf{66.4$\pm$1.3} & \textbf{68.8$\pm$1.3} & \textbf{59.6$\pm$1.2} & \textbf{81.0$\pm$0.4} & \textbf{59.9$\pm$1.7} & \textbf{80.7$\pm$1.2} \\
\bottomrule
\end{tabular}
}
\caption{Graph classification results (in \%) under node and edge density domain shifts on the Mutagenicity dataset, and feature domain shifts on DD, PROTEINS, BZR, BZR\_MD, COX2, and COX2\_MD. For convenience, PROTEINS, DD, COX2, COX2\_MD, BZR, and BZR\_MD are abbreviated as P, D, C, CM, B, and BM, respectively. \textbf{Bold} results indicate the best performance.}
\label{tab:main_results}
\vspace{-0.2cm}
\end{table*}

%% file: code/6_conclusion.tex

\section{Conclusion}

In this paper, we study semantic resolution shift in graph domain adaptation and propose \method{}, which learns comparable resolution-specific representations, prototype-guided source-to-target resolution correspondence, and posterior-conditioned target grafting. Experiments across different shifts demonstrate the effectiveness of \method{} . In future work, we plan to extend \method{} to continuous and instance-adaptive resolution spaces and other domain tasks.